\documentclass[11pt,a4paper]{article}
\usepackage[hyperref]{acl2017}
\usepackage{times}
\usepackage{latexsym}
\usepackage{enumitem}
\usepackage{url}
\usepackage{array}
\usepackage{graphicx}
\usepackage{tabularx}
\usepackage{multicol}
\usepackage{multirow}
\usepackage{amsmath}
\usepackage{float}

\aclfinalcopy

 \title{MIT at SemEval-2017 Task 10: \\Relation Extraction with Convolutional Neural Networks}
  
\author{Ji Young Lee\thanks{\hspace{3mm}These authors contributed equally to this work.}\\
	    MIT\\
	    {\tt jjylee@mit.edu}
	   \And
	Franck Dernoncourt\footnotemark[1]\\
   	MIT\\
   {\tt francky@mit.edu}
   \And
	 Peter Szolovits \\
   	MIT\\
   {\tt psz@mit.edu}
   }

\date{}

\begin{document}
\vspace{-0.1cm}
\maketitle
\begin{abstract}\vspace{-0.3cm}
Over 50 million scholarly articles have been published: they constitute a unique repository of knowledge. In particular, one may infer from them relations between scientific concepts, such as synonyms and hyponyms. Artificial neural networks have been recently explored for relation extraction. In this work, we continue this line of work and present a system based on a convolutional neural network to extract relations. Our model ranked first in the SemEval-2017 task 10 (ScienceIE
) for relation extraction in scientific articles (subtask~C).
\end{abstract}
\vspace{-0.3cm}
\section{Introduction and related work}
\vspace{-0.3cm}

The number of articles published every year keeps increasing~\cite{druss2005growth,larsen2010rate} and well over 50 million scholarly articles have been published so far~\cite{jinha2010article}.
While this repository of human knowledge contains invaluable information, it has become increasingly difficult to take advantage of all available information due to its sheer amount.

One challenge is that the knowledge present in scholarly articles is mostly unstructured. One approach to organize this knowledge is to classify each sentence~\cite{kim2011automatic,amini2012overview,hassanzadeh2014identifying,dernoncourt2016neural}. Another approach is to extract entities and relations between them, which is the focus of the ScienceIE  shared task at SemEval-2017~\cite{augenstein2017scienceie}.

Relation extraction can be seen as a process comprising two steps that can be done jointly~\cite{li2014incremental} or separately: first, entities of interest need to be identified, then the relation among each possible set of entities has to be determined. In this work, we concentrate on the second step (often referred to as relation extraction or classification) and on binary relations, i.e. relations between two entities. Extracted relations can be used for a variety of tasks such as
question-answering systems~\cite{ravichandran2002learning}, ontology extension~\cite{schutz2005relext}, and clinical trials~\cite{frunza2011extracting}.

In this paper, we describe the system that we submitted for the ScienceIE shared task. Our system is based on convolutional neural networks and ranked first for relation extraction (subtask C).

\begin{figure*}[!ht]
  \centering
  \vspace{-0.3cm}
  \includegraphics[width=\textwidth]{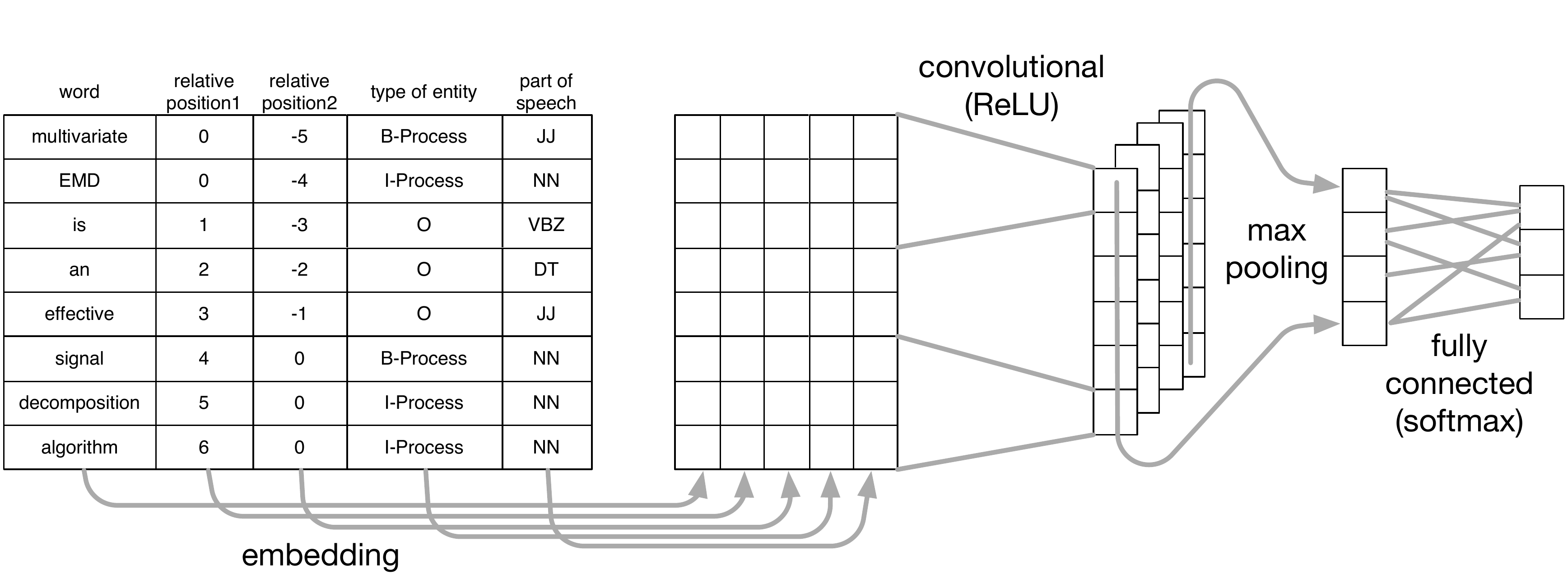}
  \vspace{-0.7cm}
  \caption{CNN architecture for relation extraction. The left table shows an example of input to the model.}
  \label{fig:model}
\end{figure*}

\begin{table*}[t]
\vspace{0.5cm}
\footnotesize
\centering
\setlength{\extrarowheight}{3pt}
\setlength{\arraycolsep}{5pt}
\begin{tabular}{|p{0.41\textwidth}|p{0.13\textwidth}|p{0.38\textwidth}|}
\hline
\textbf{Examples} & \textbf{Rule format} & \textbf{Relations detected}  \\
\hline
 transmission electron microscopy (TEM)  & A (B)	& If B is an abbreviation of A, then A and B are synonyms.  \\
 high purity standard metals (Sn, Pb, Zn, Al, Ag, Ni)  & A (B, C, ... , D)	& If any of B, C, ... , D is a hyponym of A, then all of them are hyponyms of A. \\
 (TEMs), scanning electron microscopes  & (A) B			& A and B have no relation. \\
 DOTMA/DOPE  & A/B		& A and B have no relation. \\
\hline
\end{tabular}
\vspace{-0.2cm}
\caption{Rules used for postprocessing. }\label{tab:rules}
\end{table*}

Existing systems for relation extraction can be classified into five categories~\cite{zettlemoyer2013relecture}: systems based on hand-built patterns~\cite{yangarber1998nyu}, bootstrapping methods~\cite{brin1998extracting}, unsupervised methods~\cite{gonzalez2009unsupervised}, distant supervision~\cite{snow2004learning}, and supervised methods. We focus on supervised methods, as the ScienceIE shared task provides a labeled training set.

Supervised methods for relation extraction commonly employ support vector machines~\cite{uzuner2010semantic,uzuner20112010,minard2011multi,guodong2005exploring}, na{\"\i}ve Bayes~\cite{zayaraz2015concept}, maximum entropy~\cite{sun2012active}, or conditional random fields~\cite{sutton2006introduction}. 
These methods require the practitioner to handcraft features, such as surface features, lexical features, syntactic features~\cite{grouin2010caramba} or features derived from existing ontologies~\cite{rink2011automatic}. The use of kernels based on dependency trees has also been explored~\cite{bunescu2005shortest,culotta2004dependency,zhou2007tree}.

More recently, a few studies have investigated the use of artificial neural networks for relation extraction
~\cite{socher2012semantic,nguyen2015relation,hashimoto2013simple}.
 Our approach follows this line of work.

\section{Model}
Our model for relation extraction comprises three parts: preprocessing, convolutional neural network (CNN), and postprocessing.

\subsection{Preprocessing}
The preprocessing step takes as input each raw text (i.e., in ScienceIE, a paragraph of a scientific article) as well as the location of all entities present in the text, and output several examples.
Each example is represented as a list of tokens, each with four features: the relative positions of the two entity mentions, and their entity types and part-of-speech (POS) tags. Figure~\ref{fig:model} shows an example from the ScienceIE corpus in the table on the left.

Sentence and token boundaries as well as POS tags are detected using the Stanford CoreNLP toolkit~\cite{manning-EtAl:2014:P14-5}, and 
every pair of entity mentions of the same type within each sentence boundary are considered to be of a potential relation. 
We also remove any references (e.g. [1, 2]) that are irrelevant to the task, and ensure that the sentences are not too long by eliminating the tokens before the beginning of the first entity mention and after the end of the second entity mention.

\subsection{CNN architecture}
Second, the CNN takes each preprocessed sentence as input, and predicts the relation between the two entities.
The CNN architecture, illustrated in Figure~\ref{fig:model}, consists of four main layers, similar to the one used in text classification~\cite{collobert2011natural,kim2014convolutional,lee2016sequential,gehrmann2017textclassification}.
\begin{enumerate}[noitemsep,leftmargin=*]
\item the \textbf{embedding layer} converts each feature (word, relative positions 1 / 2, type of entity, and POS tag) into an embedding vector via a lookup table and concatenates them.
\item the \textbf{convolutional layer} with ​ReLU activation transforms the embeddings into feature maps by sliding filters over the tokens.
\item the \textbf{max pooling layer} takes the most effective feature in each feature map by applying the max operator.
\item the \textbf{fully connected layer} with softmax activation outputs the probability of each relation.  
\end{enumerate} 
\vspace{-0.3cm}
 \begin{table*}[t]

\footnotesize
\centering
\setlength\tabcolsep{3.0pt}
\setlength{\extrarowheight}{3pt}
\setlength{\arraycolsep}{3pt}
\begin{tabular}{|c|ccc|ccc|}
\hline 
annotation & \multicolumn{6}{c|}{A (arg1) is a Hyponym of (rel) B (arg2)} \tabularnewline
\hline 
order in text & \multicolumn{3}{c|}{... A ... B ...} & \multicolumn{3}{c|}{... B ... A ...}  \tabularnewline
\hline 
\textbf{strategy} & \textbf{rel} & \textbf{arg1} & \textbf{arg2} & \textbf{rel} & \textbf{arg1} & \textbf{arg2} \tabularnewline
\hline 
correct order & Hypo & A & B & Hypo & A & B \tabularnewline
\hline 
correct order & Hypo & A & B & Hypo & A & B \tabularnewline
\cline{2-7} 
w/ neg. smpl. & None & B & A & None & B & A \tabularnewline
\hline 
fixed order & Hypo & A & B & Hyper & B & A \tabularnewline
\hline 
\multirow{2}{*}{any order} & Hypo & A & B & Hyper & A & B \tabularnewline
\cline{2-7} 
 & Hyper & B & A & Hypo & B & A \tabularnewline
\hline
\end{tabular}
\qquad
\begin{tabular}{|c|ccc|ccc|}
\hline
annotation & \multicolumn{6}{c|}{A (arg) is a Synonym of (rel) B (arg)}  \tabularnewline
\hline 
order in text & \multicolumn{3}{c|}{... A ... B ...} & \multicolumn{3}{c|}{... B ... A ...}   \tabularnewline
\hline 
\textbf{strategy} & \textbf{rel} & \textbf{arg1} & \textbf{arg2} & \textbf{rel} & \textbf{arg1} & \textbf{arg2} \tabularnewline
\hline 
correct order & Syn & A & B & Syn & A & B \tabularnewline
\hline  
correct order & Syn & A & B & Syn & A & B \tabularnewline
\cline{2-7} 
w/ neg. smpl. & Syn & B & A & Syn & B & A \tabularnewline
\hline
fixed order & Syn & A & B & Syn & B & A \tabularnewline
\hline  
\multirow{2}{*}{any order} & Syn & A & B & Syn & B & A \tabularnewline
\cline{2-7} 
 & Syn & B & A & Syn & A & B \tabularnewline
\hline 
\end{tabular}
\caption{Argument ordering strategies. ``w/ neg. smpl.'': with negative sampling~\cite{xu-EtAl:2015:EMNLP1}, ``rel'': relation, ``arg'': argument. ``Syn'', ``Hypo'', ``Hyper'', and ``None'' refers to the ``Synonym-of'', ``Hyponym-of'', ``Hypernym-of'', and ``None' relations. 
Note that the ``Hypernym-of'' relation is the reverse of the ``Hyponym-of'' relation, introduced in addition to the relations annotated for the dataset.}
\label{tab:labeling_strategies}
\end{table*}

\vspace{-0.3cm}
\subsection{Rule-based postprocessing}
Finally, the postprocessing step uses the rules in Table~\ref{tab:rules} to correct the relations detected by the CNN, or to detect additional relations. These rules were developed from the examples in the training set, to be consistent with common sense.

\subsection{Implementation}
During training, the objective is to maximize the log probability of the correct relation type. The model is trained using stochastic gradient descent with minibatch of size 16, updating all parameters, i.e., token embeddings, feature embeddings, CNN filter weights, and fully connected layer weights, at each gradient descent step. For regularization, dropout is applied before the fully connected layer, and early stop with a patience of 10 epochs is used based on the development set.

The token embeddings are initialized using publicly available\footnote{\url{http://nlp.stanford.edu/projects/glove/}} pre-trained token embeddings, namely GloVe~\cite{pennington2014glove} trained on Wikipedia and Gigaword 5~\cite{parker2011english}.
The feature embeddings and the other parameters of the neural network are initialized randomly.

To deal with class imbalance,
we upsampled the synonym and hyponym classes by duplicating the examples in the positive classes so that the \emph{upsampling ratio}, i.e., the ratio of the number of positive examples in each class to that of the negative examples, {}is at least 0.5.
Without the upsampling, it was impossible to train the model. 

\section{Experiments}

\subsection{Dataset}

We evaluate our model on the ScienceIE dataset~\cite{augenstein2017scienceie}, which consists of 500 journal articles evenly distributed among the domains of computer science, material sciences and physics. Three types of entities are annotated:  process, task, and material. The relation between each pair of entity of the same type within a sentence are annotated as either ``Synonym-of'', ``Hyponym-of'', or ``None''.
Table~\ref{tab:datasets} shows the number of examples for each relation class.

\begin{table} [h]
\footnotesize
\centering
\setlength\tabcolsep{4.0pt}
\setlength{\extrarowheight}{3pt}
\setlength{\arraycolsep}{5pt}
\begin{tabular}{|l|c|c|c|c|c|}
\hline
\textbf{Relation} & \textbf{Train} & \textbf{Dev} & \textbf{Test} \\
\hline
Hyponym-of & 420 & 123 & 95 \\
Synonym-of & 253 & 45 & 112 \\
None & 5355 & 1240 & 1503 \\
\hline
Total & 6028 & 1408 & 1710 \\
\hline
\end{tabular}
\caption{Number of examples for each relation class in ScienceIE. ``Dev'': Development. \vspace{-0.3cm}} \label{tab:datasets}
\end{table}

\subsection{Hyperparameters}

Table~\ref{tab:hyperparameter} details the experiment ranges and choices of hyperparameters.  
The results were quite robust to the choice of hyperparameters within the specified ranges. 

\begin{table} [h]
\footnotesize
\centering
\setlength{\extrarowheight}{3pt}
\setlength{\arraycolsep}{5pt}
\begin{tabular}{|l|c|c|}
\hline
\textbf{Hyperparameter} & \textbf{Choice} 	& \textbf{Experiment range} \\
\hline
\text{Token embedding dim.}	& 100   & 50 -- 300  \\
\text{Feature embedding dim.}	& 10 	& 5 -- 50 \\
\text{CNN filter height}			& 5 	& 3 -- 15 \\
\text{Number of CNN filters}		& 200 	& 50 -- 500 \\
\text{Dropout probability}			& 0.5 	& 0 -- 1 \\
\text{Upsampling ratio}				& 3 	& 0.5 -- 5 \\
\hline
\end{tabular}
\caption{Experiment ranges and choices of hyperparameters.\protect\footnotemark[3]} 
\label{tab:hyperparameter}
\vspace{-0.2cm}
\end{table}

\subsection{Argument ordering strategies}

 \begin{table*}[t]

\footnotesize
\centering
\setlength\tabcolsep{6.0pt}
\setlength{\extrarowheight}{3pt}
\setlength{\arraycolsep}{5pt}
\begin{tabular}{|c|c|c|ccc|ccc|ccc|}
\hline 
Labels & Training & Evaluation & \multicolumn{3}{c|}{Hyponym-of} & \multicolumn{3}{c|}{Synonym-of} & \multicolumn{3}{c|}{Micro-averaged} \tabularnewline
\cline{4-12}

 used & strategy & strategy & P & R & F1 & P & R & F1 & P & R & F1 \tabularnewline
\hline 
\multirow{5}{*}{All} & correct order & any order & 0.193	& 0.101	& 0.132	& 0.782	& 0.640	& 0.703	& 0.409	& 0.245	& 0.306 \tabularnewline
 & corr. w/ n. s. & any order & 0.431	& 0.127	& 0.196	& 0.826	& 0.756	& 0.788	&  \textbf{0.638}	& 0.295	& 0.404 \tabularnewline
 & any order & any order & 0.482	& 0.197	& 0.279	& 0.784	& 0.756	& 0.769	& 0.621	& 0.346	&  0.444 \tabularnewline
 & any order & fixed order & \textbf{0.486}	& 0.195	& 0.278	& 0.773	& 0.753	& 0.763	& 0.621	& 0.345	& 0.443 \tabularnewline
 & fixed order & any order & 0.372	& 0.218	& 0.274	& 0.743	& 0.756	& 0.749	& 0.516	& 0.362	& 0.425  \tabularnewline
 & fixed order & fixed order & 0.425	& 0.213	& 0.283	& 0.803	& 0.753	& 0.777	& 0.578	&  0.358	& 0.441 \tabularnewline
\hline
\multirow{6}{*}{Hyponym} & correct order & any order & 0.108	& 0.069	& 0.084	& -	& -	& - & -	& -	& - \tabularnewline
 & corr. w/ n. s. & any order & 0.215	& 0.115	& 0.148	& -	& -	& - & -	& -	& - \tabularnewline
 & any order & any order & 0.384	& 0.246	&  0.299	& -	& -	& - & -	& -	& - \tabularnewline
 & any order & fixed order & 0.410	& 0.235	& 0.298	& -	& -	& - & -	& -	& - \tabularnewline
 & fixed order & any order & 0.385 & \textbf{0.249} &	\textbf{0.301} &	 -	& -	& - & -	& -	& -  \tabularnewline
 & fixed order & fixed order & 0.409 &	0.237 &	0.297 &	 -	& -	& - & -	& -	& -  \tabularnewline
\hline
\multirow{2}{*}{Synonym} & any order & any order & -	& -	& -	& \textbf{0.855}	& 0.771	& 0.811 & -	& -	& - \tabularnewline
 & any order & fixed order & -	& -	& -	& 0.852	&  \textbf{0.776}	&  \textbf{0.812} & -	& -	& - \tabularnewline
\hline 
Hyp+Syn & any + any & any + fixed & 0.385 &0.228 & 0.285 & \textbf{0.857} & 0.771 & \textbf{0.812} & 0.553 & \textbf{0.373} & \textbf{0.445} \tabularnewline
\hline

\end{tabular}
\caption{Results for various ordering strategies on the development set of the ScienceIE dataset, averaged over 10 runs each.\footnotemark[3] ``corr. w/ n. s.'': correct order with negative sampling. Hyp+Syn is obtained by merging the output of the best hyponym classifier and that of the best synonym classifier. 
} 
\label{tab:result_strategies}
\end{table*}

\footnotetext[3]{For these experiments, we used the official training set as the training/development set with a $75\%/25\%$ split, and the official development set as the test set.}
One of the main challenges in relation extraction is the ordering of arguments in relations, as many relations are \emph{order-sensitive}.
For example, consider the sentence ``A dog is an animal." If we set ``dog'' be the first argument and ``animal'' the second, then the corresponding relation is ``Hyponym-of''; however, if we reverse the argument order,
then the ``Hyponym-of'' relation does not hold any more.   

Therefore, it is crucial to ensure that 1) the CNN is provided with the information about the argument order, and 2) it is able to utilize the given information efficiently. 
In our work, the former point is addressed by providing the CNN with the two relative position features compared to the first and the second argument of the relation respectively. In order to certify the latter point, we experimented with four strategies for argument ordering, outlined in Table~\ref{tab:labeling_strategies}.

\section{Results and Discussion}

Table~\ref{tab:result_strategies} shows the results from experimenting with various argument ordering strategies. The correct order strategy performed the worst, but the negative sampling improved over it slightly, while the fixed order and any order strategies performed the best. The latter two strategies performed almost equally well in terms of micro-averaged F1-score. This implies that for relation extraction it may be advantageous to use both the original relation classes as well as their ``reverse'' relation classes for training, instead of using only the original relation classes with the ``correct'' argument ordering (with or without the negative sampling). Moreover, ordering the argument as the order of appearance in the text and training once per relation (i.e., fixed order) is as efficient as training each relation as two examples in two possible argument ordering, one with the original relation class and the other with the reverse relation class (i.e., any order), despite the small size of the dataset.

The difference in performance between the correct order versus the fixed or any order strategies is more prominent for the ``Hyponym-of'' relation than for the ``Synonym-of'' relation. This is expected, since the argument ordering strategy is different only for the order-sensitive ``Hyponym-of'' relation. It is somewhat surprising though, that the correct order strategy performs worse then the other strategies even for order-insensitive ``Synonym-of'' relation. 
This may be due to the fact that the model
does not see any training examples with the reversed argument ordering for the ``Synonym-of'' relation. 
In comparison, the negative sampling strategy, which learns from both the original and reversed argument ordering for the ``Synonym-of'' relation, the performance is comparable to the two best performing strategies.

We have also experimented with different evaluation strategies for the models trained with the any order and fixed order strategies. When the model is trained with the any order strategy, the choice of the evaluation strategy does not impact the performance. In contrast, when the model is trained with the fixed order strategy, it performs better if the same strategy is used for evaluation. This may be the reason that the model trained with the correct order strategy does not perform as well, since it has to be evaluated with a different strategy from training, namely the any order strategy, as we do not know the correct ordering of arguments for examples in the test set. 

We have also tried training binary classifiers for the ``Hyponym-of'' and the ``Synonym-of'' relations separately and then merging the outputs of the best classifiers for each relations. While the binary classifiers individually performed better than the multi-way classifier for each corresponding relation class, the overall performance based on the micro-averaged F1-score did not improve over the multi-way classifier after merging the outputs of the hyponym and the synonym classifiers.

Based on the results from the argument ordering strategy experiments, we submitted the model trained using the fixed order strategy, which ranked number one in the challenge. The result is shown in Table~\ref{tab:result}.

\begin{table} [h]
\footnotesize
\centering
\setlength\tabcolsep{4.0pt}
\setlength{\extrarowheight}{3pt}
\setlength{\arraycolsep}{5pt}
\begin{tabular}{|l|c|c|c|c|c|}
\hline
\textbf{Relation} & \textbf{Precision} & \textbf{Recall} & \textbf{F1-score} \\
\hline
Synonym-of & 0.820 & 0.813 & 0.816\\
Hyponym-of & 0.455 & 0.421 & 0.437\\
\hline
Micro-averaged & 0.658 & 0.633 & 0.645\\
\hline
\end{tabular}
\caption{Result on the test set of the ScienceIE dataset, using the official train/dev/test split.} \label{tab:result}
\end{table}

\begin{figure}[h]
  \centering
  \vspace{-0.3cm}
  \includegraphics[width=0.45\textwidth]{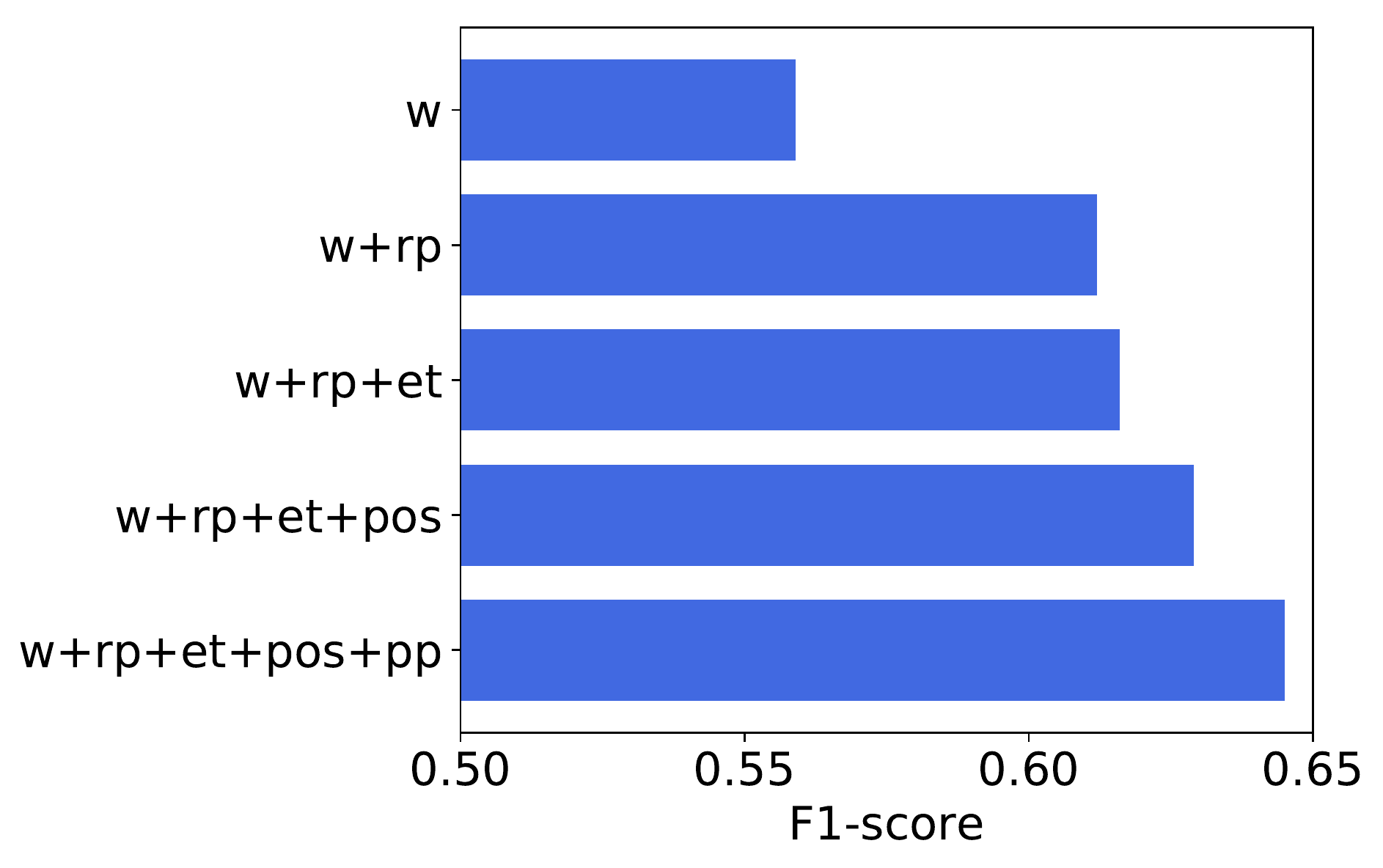}
  \vspace{-0.3cm}
  \caption{Importance of features of CNN and postprocessing rules. w: word embeddings, rp: relative positions to the first and the second arguments, et: entity types, pos: POS tags.}
  \label{fig:ablation1}
\end{figure}

To quantify the importance of various features of our model, we trained the model by gradually adding more features one by one, from word embeddings, relative positions, and entity types to POS tags in order. The results on the importance of the features as well as postprocessing are shown in Figure~\ref{fig:ablation1}. Adding the relative position features improved the performance the most, while adding the entity type improved it the least.

\begin{figure}[h]
  \centering

  \includegraphics[width=0.45\textwidth]{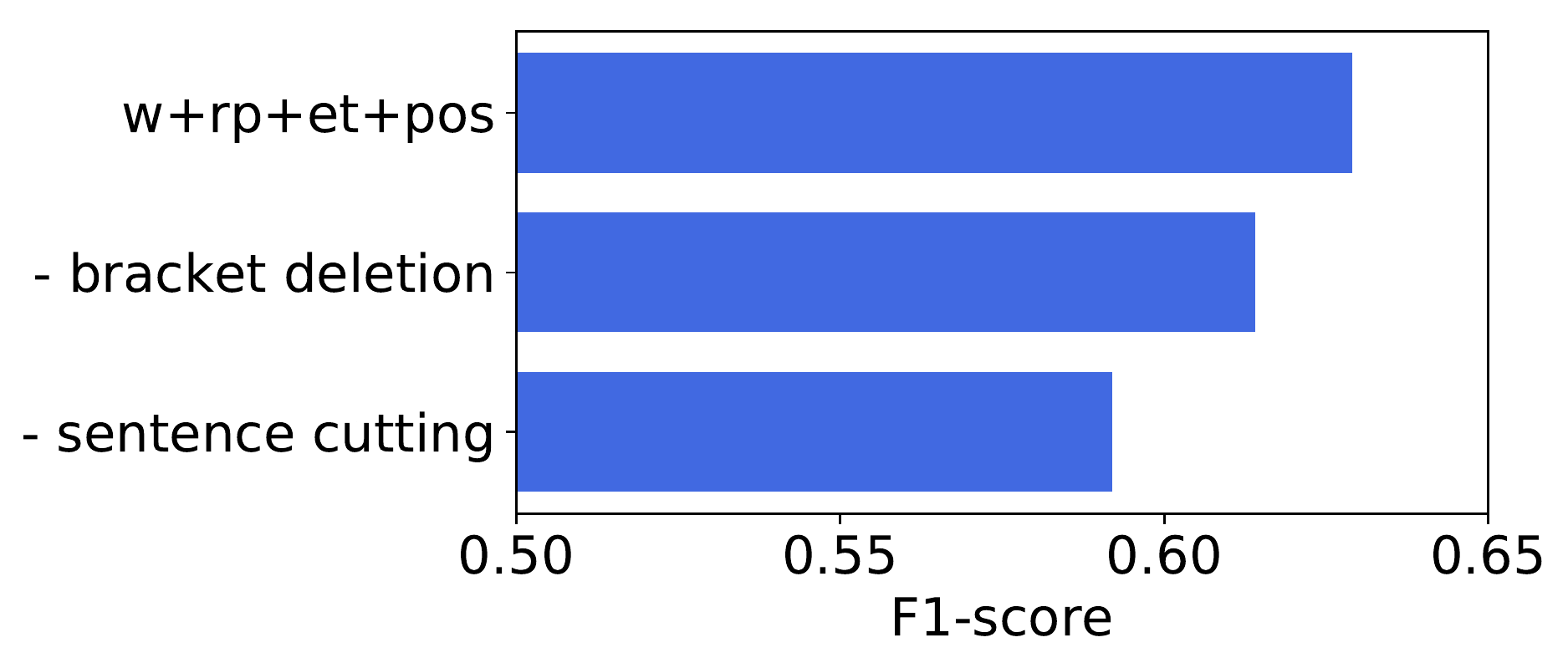}
  \vspace{-0.3cm}
  \caption{Impact of bracket deletion and sentence cutting. ``w+rp+et+pos'' represents the CNN model trained using all features with both bracket deletion and sentence cutting during preprocessing. ``-bracket deletion'' is the same model trained only without bracket deletion, and ``-sentence cutting'' just without sentence cutting.}
  \label{fig:ablation2}
\end{figure}

Figure~\ref{fig:ablation2} quantifies the impact of two preprocessing steps, deleting brackets and cutting sentences, introduced to compensate for the small dataset size. Cutting the sentence before the first entity and after the second entity resulted in a dramatic impact on the performance, while deleting brackets (i.e., removing the reference marks) improve the performance modestly. This implies that the text between the two entities contains most of the information about the relation between them.

\vspace{-0.5cm}

\section{Conclusion}

In this article we have presented an ANN-based approach to relation extraction, which ranked first in the SemEval-2017 task 10 (ScienceIE) for relation extraction in scientific articles (subtask C). We have experimented with various strategies to incorporate argument ordering for ordering-sensitive relations, showing that an efficient strategy is to fix the arguments ordering as appears on the text by introducing reverse relations. We have also demonstrated that cutting the sentence before the first entity and after the second entity is effective for small datasets.

\section*{Acknowledgments}

The authors would like to thank the Science IE organizers for
managing this shared task. 
The project was supported by Philips Research. The content is solely the responsibility of the authors and does not necessarily represent the official views of Philips Research.

\bibliography{acl2017}
\bibliographystyle{acl_natbib}

\end{document}